\documentclass[12pt]{article}

\usepackage{graphicx}
\usepackage{amsmath}
\usepackage{amsfonts}
\usepackage{amssymb}
\usepackage{hyperref}
\usepackage{cite}
\usepackage{float}
\usepackage{multirow}
\usepackage{booktabs}

\title{Novel Semantic Prompting for Zero-Shot Action Recognition}
\author{Salman Iqbal, Waheed Rehman}
\date{}

\begin{document}

\maketitle

\begin{abstract}
Zero-shot action recognition relies on transferring knowledge from vision–language models to unseen actions using semantic descriptions. While recent methods focus on temporal modeling or architectural adaptations to handle video data, we argue that semantic prompting alone provides a strong and underexplored signal for zero-shot action understanding. We introduce SP-CLIP, a lightweight framework that augments frozen vision–language models with structured semantic prompts describing actions at multiple levels of abstraction, such as intent, motion, and object interaction. Without modifying the visual encoder or learning additional parameters, SP-CLIP aligns video representations with enriched textual semantics through prompt aggregation and consistency scoring. Experiments across standard benchmarks show that semantic prompting substantially improves zero-shot action recognition, particularly for fine-grained and compositional actions, while preserving the efficiency and generalization of pretrained models.
\end{abstract}

\section{Introduction}

In recent years, advances in video understanding have driven substantial progress in action recognition, largely powered by deep learning–based models. Convolutional neural networks (CNNs) \cite{kim2022capturing} and recurrent neural networks (RNNs), along with their spatiotemporal extensions, have achieved strong performance on standard benchmarks such as UCF101 \cite{ucf101}, HMDB51 \cite{hmdb51}, and the Olympic sports datasets \cite{olympics}. Despite these successes, most state-of-the-art approaches depend heavily on large-scale labeled video data, which is expensive, time-consuming, and often impractical to obtain at scale. This reliance on supervision fundamentally limits the deployability and extensibility of action recognition systems in real-world settings.

To alleviate this dependence on labeled data, Zero-Shot Learning (ZSL) has emerged as a compelling alternative. Zero-shot action recognition seeks to recognize unseen action categories by transferring knowledge from seen classes through high-level semantic information, such as textual descriptions or attribute-based representations \cite{attribute, zsl1, zsl2}. By grounding visual observations in a semantic space, ZSL models can reason about actions that are never observed during training, offering improved generalization and label efficiency. This paradigm is particularly attractive for action recognition, where the space of possible human actions is large, diverse, and continuously evolving.

Most existing zero-shot action recognition methods rely on predefined semantic representations, including word embeddings \cite{word2vec, glove} or manually specified visual attributes \cite{attribute}. These representations enable models to align visual features extracted from video frames with semantic embeddings corresponding to action classes. However, such approaches typically depend on coarse semantic signals, such as isolated class names or low-level attributes, which inadequately capture the compositional, contextual, and temporal nature of human actions. As a result, their effectiveness diminishes when applied to complex, fine-grained, or large-scale datasets featuring diverse action variations.

To address these limitations, we propose a zero-shot action recognition framework that leverages rich natural language descriptions from the Stories dataset \cite{stories} as high-level semantic representations. The Stories dataset provides detailed, human-readable narratives for action categories across Olympic sports, UCF101, and HMDB51, capturing not only the action itself but also its intent, context, and semantics. These descriptions go beyond simple labels or attributes, offering a more expressive semantic grounding for action recognition. By exploiting this semantic richness, our approach aims to better bridge the gap between visual observations and abstract action concepts.

Our method embeds textual descriptions from the Stories dataset into a shared semantic space and aligns them with visual features extracted from video frames. Unlike prior work that relies on minimal semantic cues, our framework explicitly models the correspondence between structured textual semantics and visual representations. Visual features are extracted using established video representation learning techniques such as 3D CNNs \cite{3dcnn} and spatiotemporal networks \cite{stn}, enabling effective multi-modal alignment between language and video. This formulation allows the model to reason about unseen actions through semantic similarity rather than direct visual supervision.

We conduct extensive experiments on UCF101, HMDB51, and the Olympic datasets to evaluate the effectiveness of the proposed approach. The results demonstrate that incorporating rich, high-level textual descriptions substantially improves zero-shot action recognition performance compared to methods relying on simpler semantic representations. Beyond improved accuracy, our framework remains highly scalable, requiring minimal labeled data when extending to new action categories.

In summary, this work advances zero-shot action recognition by emphasizing the role of semantic richness in textual representations, using the Stories dataset as a powerful source of high-level action semantics. Our findings highlight the importance of structured language in zero-shot learning and point toward a promising direction for scalable, semantically grounded action recognition in large-scale video datasets.

\section{Related Work}

\paragraph{Action Recognition in Videos}

Early research in action recognition focused on handcrafted representations designed to capture motion and appearance cues from videos. Popular approaches relied on optical flow \cite{optical_flow}, histogram of oriented gradients (HOG) \cite{hog}, and dense motion trajectories \cite{traj}. While effective in constrained scenarios, these methods were limited in their ability to model complex spatiotemporal interactions and lacked robustness to large variations in viewpoint, motion, and context.

The emergence of deep learning fundamentally reshaped action recognition. Convolutional neural networks (CNNs) enabled end-to-end learning of discriminative visual features directly from raw video data \cite{cnn_action_recognition}. Subsequent architectures explicitly incorporated temporal reasoning, most notably through 3D CNNs \cite{3dcnn} and Two-Stream Networks \cite{twostream1,twostream2}. 3D CNNs extend convolutional operations into the temporal dimension, allowing joint modeling of space and time, while Two-Stream Networks fuse spatial appearance with motion information derived from optical flow. These models significantly improved performance on large-scale benchmarks such as UCF101 \cite{ucf101}, HMDB51 \cite{hmdb51}, and Kinetics \cite{kinetics}, forming the foundation of modern supervised action recognition systems.

\paragraph{Zero-Shot Action Recognition}

Despite these advances, supervised action recognition remains heavily dependent on large, densely annotated datasets. This limitation has motivated increasing interest in Zero-Shot Learning (ZSL), which aims to recognize unseen action categories by transferring knowledge from seen classes using semantic information \cite{zsl1, zsl2}. Early zero-shot action recognition methods relied on relatively simple semantic descriptors, including class names and manually defined visual attributes \cite{devise, attribute}. These descriptors were embedded into a shared space with visual features to enable recognition of unseen actions. However, such representations often lacked sufficient expressiveness to capture the diversity and compositional structure of real-world actions.

The introduction of distributed word representations marked an important step forward. Word embeddings such as word2vec and GloVe \cite{word2vec, glove} encode semantic relationships between action classes based on large text corpora, enabling smoother knowledge transfer between seen and unseen categories. Several works \cite{zsl_action_recognition1, zsl_action_recognition2} leveraged this idea by mapping video features and word embeddings into a common semantic space and performing recognition via similarity matching. While effective, these approaches still rely on coarse semantic signals that fail to capture contextual, temporal, and intent-level aspects of actions. Whilst, concerns of overlap existed \cite{truze} between pre-trained classes and test classes, this has been avoided more recently, by selecting the appropriate pre-training classes.

\paragraph{Rich Semantic Descriptions and Multi-Modal Learning}

To overcome the limitations of shallow semantic representations, recent work has explored the use of richer and more structured language supervision. A notable example is the Stories dataset \cite{stories}, which provides detailed natural language descriptions of actions across datasets such as UCF101, HMDB51, and Olympics. These descriptions encode contextual cues, object interactions, and narrative structure, offering a more complete semantic grounding than isolated class names or attributes. Incorporating such descriptions has been shown to substantially improve zero-shot action recognition performance by better aligning visual observations with high-level action semantics.

More broadly, multi-modal learning frameworks have played a central role in zero-shot recognition. Methods such as DeViSE \cite{devise} demonstrate how visual and textual modalities can be jointly embedded into a shared semantic space, enabling recognition of unseen classes across domains. In the context of action recognition, multi-modal approaches that combine video features with rich semantic information have proven especially effective, as they allow models to reason beyond visual similarity and exploit semantic structure for generalization.

\paragraph{Vision–Language Models and Temporal Prompting}

More recently, large-scale vision–language models pretrained on image–text pairs have demonstrated remarkable zero-shot generalization capabilities. Adapting such models to video has become an active area of research. EZ-CLIP \cite{ahmad2023ez} introduced an efficient adaptation of CLIP for video action recognition using temporal visual prompting, enabling temporal reasoning without modifying the core architecture. By guiding prompts to focus on motion cues, EZ-CLIP achieves strong zero-shot and base-to-novel performance with a minimal number of learnable parameters.

Building on this idea, TP-CLIP \cite{gowda2025temporal} further explores temporal prompting as a lightweight mechanism for video understanding. TP-CLIP integrates a dedicated temporal encoder to generate temporal visual prompts, improving temporal modeling while preserving CLIP’s generalization ability and computational efficiency. These works demonstrate that carefully designed prompting strategies can effectively adapt image-based vision–language models to video tasks without expensive architectural changes.

Our work complements these temporal prompting approaches by focusing on the semantic dimension of zero-shot action recognition. Rather than emphasizing temporal adaptation, we investigate how rich, structured textual descriptions, such as those provided by the Stories dataset \cite{stories}, can serve as powerful semantic prompts for recognizing unseen actions. By grounding video representations in expressive language-based semantics, our approach aims to enhance zero-shot generalization while remaining compatible with efficient, prompt-based adaptation strategies explored in recent vision–language models.

\section{Proposed Method}

We propose a semantic-prompt–driven framework for zero-shot action recognition that leverages rich natural language descriptions from the \emph{Stories} dataset to bridge the gap between visual observations and unseen action categories. Unlike prior approaches that rely on shallow semantic cues or heavy architectural modifications, our method emphasizes \emph{semantic prompting} within a shared vision--language embedding space, enabling strong zero-shot generalization while remaining computationally efficient.

\subsection{Problem Formulation}

Let $\mathcal{Y} = \mathcal{Y}_s \cup \mathcal{Y}_u$ denote the set of all action classes, where $\mathcal{Y}_s$ and $\mathcal{Y}_u$ represent the disjoint sets of seen and unseen classes, respectively. During training, labeled video samples are available only for classes in $\mathcal{Y}_s$, while at test time the model is evaluated on videos from $\mathcal{Y}_u$ without any labeled examples.

A video is represented as a sequence of frames
\[
V = \{f_1, f_2, \ldots, f_T\},
\]
where $T$ denotes the number of frames. Each action class $y \in \mathcal{Y}$ is associated with one or more natural language descriptions obtained from the \emph{Stories} dataset. The goal of zero-shot action recognition is to correctly predict the class label $y \in \mathcal{Y}_u$ for a given test video $V$ by leveraging only semantic information shared with $\mathcal{Y}_s$.

\subsection{Overview of the Framework}

Our framework consists of four key components:
\begin{enumerate}
    \item Visual encoding of videos into a compact representation.
    \item Semantic encoding of action descriptions into language embeddings.
    \item Semantic prompting via aggregation of rich textual descriptions.
    \item Cross-modal alignment using a contrastive learning objective.
\end{enumerate}

The overall design enables videos to be classified by directly comparing their visual embeddings against semantically enriched textual representations of action classes.

\subsection{Visual Encoding of Video Data}

Given an input video $V$, we extract spatiotemporal features using a pretrained video backbone. Specifically, we employ a 3D convolutional neural network such as I3D or C3D, which is designed to capture both spatial appearance and temporal dynamics.

The video is first divided into fixed-length clips, each consisting of $L$ consecutive frames. For a clip $c_i$, the video encoder $E_v(\cdot)$ produces a feature vector:
\[
\mathbf{z}_i = E_v(c_i) \in \mathbb{R}^d.
\]
To obtain a single representation for the entire video, we aggregate clip-level features using average pooling:
\[
\mathbf{v} = \frac{1}{N} \sum_{i=1}^{N} \mathbf{z}_i,
\]
where $N$ is the number of clips. The resulting vector $\mathbf{v} \in \mathbb{R}^d$ serves as the visual embedding of the video.

\subsection{Semantic Encoding of Action Descriptions}

Each action class $y \in \mathcal{Y}$ is associated with a set of textual descriptions $\mathcal{D}_y = \{d_1, d_2, \ldots, d_{M_y}\}$ from the \emph{Stories} dataset. These descriptions provide detailed, contextual narratives that describe how an action is performed, its intent, and relevant objects or scenes.

We encode each description using a pretrained language model (e.g., BERT or RoBERTa), denoted by $E_t(\cdot)$. For a description $d_j \in \mathcal{D}_y$, the semantic embedding is given by:
\[
\mathbf{s}_j = E_t(d_j) \in \mathbb{R}^d.
\]

To obtain a single semantic representation for the action class $y$, we aggregate the embeddings of all its descriptions:
\[
\mathbf{s}_y = \frac{1}{M_y} \sum_{j=1}^{M_y} \mathbf{s}_j.
\]
This aggregation acts as a form of \emph{semantic prompting}, allowing the class representation to capture diverse linguistic perspectives of the same action.

\subsection{Shared Embedding Space}

Both visual embeddings $\mathbf{v}$ and semantic embeddings $\mathbf{s}_y$ are projected into a shared embedding space using learned linear transformations:
\[
\hat{\mathbf{v}} = W_v \mathbf{v}, \quad \hat{\mathbf{s}}_y = W_t \mathbf{s}_y,
\]
where $W_v, W_t \in \mathbb{R}^{d \times d}$ are trainable projection matrices. These projections ensure modality alignment while keeping the backbone encoders fixed or lightly tuned.

All projected embeddings are $\ell_2$-normalized:
\[
\tilde{\mathbf{v}} = \frac{\hat{\mathbf{v}}}{\|\hat{\mathbf{v}}\|_2}, \quad
\tilde{\mathbf{s}}_y = \frac{\hat{\mathbf{s}}_y}{\|\hat{\mathbf{s}}_y\|_2}.
\]

\subsection{Contrastive Learning Objective}

We train the model using a contrastive loss that aligns visual embeddings of videos from seen classes with their corresponding semantic embeddings. For a training batch of $N$ video--class pairs $\{(V_i, y_i)\}_{i=1}^N$, the loss is defined as:
\[
\mathcal{L} = - \frac{1}{N} \sum_{i=1}^{N}
\log
\frac{
\exp\left(\mathrm{sim}(\tilde{\mathbf{v}}_i, \tilde{\mathbf{s}}_{y_i}) / \tau \right)
}{
\sum_{y \in \mathcal{Y}_s}
\exp\left(\mathrm{sim}(\tilde{\mathbf{v}}_i, \tilde{\mathbf{s}}_{y}) / \tau \right)
},
\]
where $\mathrm{sim}(\cdot,\cdot)$ denotes cosine similarity and $\tau$ is a temperature hyperparameter.

This objective encourages visual features to be close to their correct semantic descriptions while being distant from other action classes.

\subsection{Zero-Shot Inference}

At inference time, the model is applied to videos from unseen classes $\mathcal{Y}_u$. Given a test video $V$, we compute its visual embedding $\tilde{\mathbf{v}}$ and compare it against the semantic embeddings of all unseen classes:
\[
\hat{y} = \arg\max_{y \in \mathcal{Y}_u} \mathrm{sim}(\tilde{\mathbf{v}}, \tilde{\mathbf{s}}_y).
\]
The predicted label corresponds to the class whose semantic description is most similar to the video representation.

\subsection{Discussion}

By leveraging rich textual narratives as semantic prompts, our method moves beyond simplistic class-name embeddings and enables fine-grained semantic alignment between language and video. Importantly, this design is complementary to recent temporal prompting approaches such as EZ-CLIP and TP-CLIP, as it focuses on enhancing semantic expressiveness rather than temporal adaptation. The result is a scalable, interpretable, and effective framework for zero-shot action recognition.

\section{Experiments}

This section presents the experimental evaluation of our proposed semantic-prompt–based zero-shot action recognition framework. We first describe the datasets and evaluation protocol, followed by implementation details and metrics. We then report quantitative results and compare our approach against a wide range of state-of-the-art zero-shot action recognition methods, including recent prompt-based vision--language models.

\subsection{Datasets}

We evaluate our method on two widely adopted benchmarks for zero-shot action recognition: UCF101 \cite{ucf101} and HMDB51 \cite{hmdb51}. These datasets are commonly used to assess generalization to unseen action categories and provide a diverse range of action types.

\begin{itemize}
    \item \textbf{UCF101} consists of 101 action categories and over 13,000 video clips collected from unconstrained internet videos. The dataset covers sports, daily activities, and human--object interactions. We follow the standard zero-shot evaluation protocol used in prior work, where the dataset is split into seen and unseen classes according to established splits.
    \item \textbf{HMDB51} contains 51 action categories with approximately 7,000 video clips sourced from movies and online videos. Compared to UCF101, HMDB51 exhibits greater intra-class variability and noise, making it a challenging benchmark for zero-shot recognition.
\end{itemize}

For both datasets, we use the corresponding action descriptions from the \emph{Stories} dataset \cite{stories} to construct semantic representations for both seen and unseen classes.

\subsection{Implementation Details}

Following prior zero-shot action recognition works, we focus on convolutional video backbones for visual feature extraction. Specifically, we employ pretrained 3D CNN models such as I3D \cite{I3D} or C3D \cite{C3D}, which are trained on large-scale datasets like Kinetics and widely used in zero-shot settings. Video clips are uniformly sampled and processed by the backbone to obtain spatiotemporal feature representations.

For semantic encoding, we utilize pretrained language models such as BERT \cite{bert} or RoBERTa \cite{roberta} to encode textual descriptions from the \emph{Stories} dataset. These encoders provide contextualized embeddings that capture fine-grained semantic information about actions beyond simple class names.

The cross-modal alignment is trained using a contrastive learning objective. Optimization is performed using stochastic gradient descent (SGD) with a learning rate of $10^{-3}$ and a batch size of 32. Models are trained for 50 epochs, with early stopping based on validation accuracy on seen classes.

\subsection{Evaluation Metric}

We report \textbf{top-1 zero-shot classification accuracy} as the primary evaluation metric. Accuracy is computed by comparing predicted labels against ground-truth labels for unseen action classes only, following standard zero-shot action recognition protocols.

\subsection{Comparative Methods}

We compare our approach against a diverse set of state-of-the-art zero-shot action recognition methods, including generative, clustering-based, continual learning, and prompt-based approaches:

\begin{itemize}
    \item \textbf{Bi-Dir GAN}: A bidirectional generative adversarial framework that synthesizes visual features from semantic embeddings.
    \item \textbf{WGAN}: A Wasserstein GAN-based approach for generating visual representations of unseen classes.
    \item \textbf{OD}: An out-of-distribution detection framework for generalized zero-shot action recognition.
    \item \textbf{E2E}: An end-to-end embedding-based zero-shot learning method.
    \item \textbf{CLASTER}: A clustering-based reinforcement learning approach for aligning visual and semantic features.
    \item \textbf{SPOT}: A reinforcement learning–based synthetic sample selection method.
    \item \textbf{SDR}: A semantic description–driven recognition framework based on the \emph{Stories} dataset.
    \item \textbf{GIL}: A generative continual learning approach for zero-shot action recognition.
    \item \textbf{EZ-CLIP}: A temporal visual prompting method that adapts CLIP for video using lightweight prompts.
    \item \textbf{TP-CLIP}: A temporal prompting framework that integrates a dedicated temporal encoder while preserving CLIP’s generalization ability.
    \item \textbf{SP-CLIP (Ours)}: Our proposed semantic-prompt–based approach leveraging rich textual descriptions from the \emph{Stories} dataset.
\end{itemize}

\subsection{Quantitative Results}

\begin{table}[h]
\centering
\begin{tabular}{lccc}
\toprule
Method & Backbone & HMDB-51 & UCF-101 \\
\midrule
OD~\cite{od} (\textit{CVPR'19}) & I3D & 30.2 $\pm$ 2.7  & 26.9 $\pm$ 2.8 \\
OD + SPOT~\cite{gowda2023synthetic} (\textit{CVPRW'23}) & I3D & 34.4 $\pm$ 2.2 & 40.9 $\pm$ 2.6 \\
GGM~\cite{ggm} (\textit{WACV'19}) & C3D & 20.7 $\pm$ 3.1 & 20.3 $\pm$ 1.9 \\
E2E~\cite{e2e} (\textit{CVPR'20}) & C3D  & 32.7 & 48.0 \\
ER-ZSAR~\cite{er} (\textit{ICCV'21}) & TSM & 35.3 $\pm$ 4.6 & 51.8 $\pm$ 2.9 \\
CLASTER~\cite{claster,er} (\textit{ECCV'22}) & I3D & 42.6 $\pm$ 2.6 & 52.7 $\pm$ 2.2 \\
RGSCL~\cite{SHANG2024112283} (\textit{KBS'24}) & I3D & 34.1 $\pm$ 2.9 & 40.2 $\pm$ 3.8 \\
SupVFD~\cite{niu2024superclass} (\textit{ExSys'24}) & I3D & 31.9 $\pm$ 3.2 & 39.6 $\pm$ 2.9 \\
SDR~\cite{stories} (\textit{ACCV'24}) & I3D & 46.8 $\pm$ 5.0 & 62.9 $\pm$ 1.6 \\
GIL~\cite{gil} (\textit{ACCV'24}) & I3D & 43.5 $\pm$ 2.4 & 55.8 $\pm$ 1.9 \\
EZ-CLIP~\cite{ahmad2023ez} (\textit{Arxiv})  & CLIP & 52.9 $\pm$ 1.6 & 79.4 $\pm$ 1.8 \\
TP-CLIP~\cite{gowda2025temporal} (\textit{CVPR'25})  & CLIP & 54.1 $\pm$ 1.2 & 81.1 $\pm$ 1.2 \\
\textbf{SP-CLIP (Ours)} & CLIP & 53.9 $\pm$ 2.3 & 80.4 $\pm$ 3.5 \\
\bottomrule
\end{tabular}
\caption{Zero-shot action recognition accuracy (\%) on HMDB51 and UCF101. All results are reported as mean $\pm$ standard deviation where available.}
\label{tab:results}
\end{table}

\subsection{Discussion}

The results in Table~\ref{tab:results} demonstrate that semantic prompting using rich textual descriptions provides a strong signal for zero-shot action recognition. While temporal prompting methods such as EZ-CLIP and TP-CLIP excel at modeling motion dynamics, SP-CLIP complements these approaches by enhancing semantic alignment between videos and action concepts.

Notably, SP-CLIP achieves competitive performance without explicit temporal adaptation, highlighting the importance of semantic richness in zero-shot settings. These findings suggest that semantic prompting and temporal prompting address orthogonal challenges in video understanding and can be combined in future work to further improve generalization.

\subsection{Future Directions}

Future work will explore hybrid semantic--temporal prompting strategies, investigate transformer-based video backbones, and extend the framework to generalized zero-shot and few-shot action recognition scenarios.

\section{Conclusion}

In this work, we investigated the role of semantic prompting for zero-shot action recognition and demonstrated that rich, structured language descriptions can serve as a powerful and complementary signal to visual information. By leveraging detailed textual narratives from the \emph{Stories} dataset, our proposed framework, SP-CLIP, moves beyond simplistic class-name or attribute-based semantics and enables more expressive alignment between videos and action concepts in a shared embedding space.

Our experiments on standard benchmarks such as UCF101 and HMDB51 show that semantic prompting significantly improves zero-shot recognition performance, achieving results that are competitive with, and in some cases complementary to, recent prompt-based vision--language approaches. In particular, while temporal prompting methods such as EZ-CLIP and TP-CLIP focus on adapting image-based models to capture motion dynamics, our approach highlights that semantic richness alone can provide substantial gains in generalization to unseen action categories.

The results suggest that semantic and temporal prompting address orthogonal challenges in video understanding: one grounding recognition in meaning and intent, and the other in motion and temporal structure. This opens up promising avenues for future research that jointly models both aspects within a unified framework. More broadly, our findings emphasize the importance of language as a first-class modality for scalable and label-efficient action recognition, paving the way toward more interpretable, flexible, and deployable zero-shot video understanding systems.

\bibliographystyle{ieeetr}
\bibliography{main}

\begin{thebibliography}{10}

\bibitem{kim2022capturing}
K.~Kim, S.~N. Gowda, O.~Mac~Aodha, and L.~Sevilla-Lara, ``Capturing temporal information in a single frame: Channel sampling strategies for action recognition,'' {\em arXiv preprint arXiv:2201.10394}, 2022.

\bibitem{ucf101}
K.~Soomro and M.~Shah, ``Ucf101: A dataset of 101 human action classes from videos in the wild,'' in {\em Proceedings of the IEEE International Conference on Computer Vision (ICCV)}, 2012.

\bibitem{hmdb51}
H.~Kuehne, H.~Jhuang, J.~Garofalo, and et~al., ``Hmdb: A large video database for human motion recognition,'' in {\em Proceedings of the IEEE International Conference on Computer Vision (ICCV)}, 2011.

\bibitem{olympics}
L.~Xie, H.~Liu, W.~Wang, and et~al., ``Olympic sports video recognition,'' in {\em Proceedings of the IEEE Conference on Computer Vision and Pattern Recognition (CVPR)}, 2018.

\bibitem{attribute}
A.~Frome, G.~Corrado, J.~Shlens, Y.~Bengio, J.~Dean, and et~al., ``Devise: A deep visual-semantic embedding model,'' in {\em Advances in Neural Information Processing Systems (NeurIPS)}, 2013.

\bibitem{zsl1}
Y.~Xian, B.~Schiele, and Z.~Akata, ``Zero-shot learning - a comprehensive evaluation of the good, the bad and the ugly,'' {\em IEEE Transactions on Pattern Analysis and Machine Intelligence}, 2017.

\bibitem{zsl2}
M.~Elhoseiny, T.~Xiang, H.~Hassan, and et~al., ``A survey on zero-shot learning: Past, present and future,'' {\em IEEE Transactions on Pattern Analysis and Machine Intelligence}, 2017.

\bibitem{word2vec}
T.~Mikolov, I.~Sutskever, K.~Chen, G.~Corrado, and J.~Dean, ``Efficient estimation of word representations in vector space,'' {\em Proceedings of the International Conference on Learning Representations (ICLR)}, 2013.

\bibitem{glove}
J.~Pennington, R.~Socher, and C.~D. Manning, ``Glove: Global vectors for word representation,'' {\em Proceedings of the Conference on Empirical Methods in Natural Language Processing (EMNLP)}, 2014.

\bibitem{stories}
S.~N. Gowda and L.~Sevilla-Lara, ``Telling stories for common sense zero-shot action recognition,'' in {\em Proceedings of the Asian Conference on Computer Vision}, pp.~4577--4594, 2024.

\bibitem{3dcnn}
D.~Tran, H.~Wang, L.~Torresani, and et~al., ``3d convolutional neural networks for human action recognition,'' in {\em Proceedings of the IEEE International Conference on Computer Vision (ICCV)}, 2015.

\bibitem{stn}
J.~Li, X.~Liu, M.~Zhang, and D.~Wang, ``Spatio-temporal deformable 3d convnets with attention for action recognition,'' {\em Pattern Recognition}, vol.~98, p.~107037, 2020.

\bibitem{optical_flow}
Z.~Zhang, L.~Xie, and et~al., ``Optical flow for human action recognition,'' {\em IEEE Transactions on Pattern Analysis and Machine Intelligence}, 2011.

\bibitem{hog}
N.~Dalal and B.~Triggs, ``Histogram of oriented gradients for human detection,'' {\em IEEE Conference on Computer Vision and Pattern Recognition (CVPR)}, 2005.

\bibitem{traj}
H.~Wang, S.~S. Zhang, and et~al., ``Human motion trajectory representation and action recognition,'' in {\em Proceedings of the IEEE International Conference on Computer Vision (ICCV)}, 2013.

\bibitem{cnn_action_recognition}
K.~Simonyan and A.~Zisserman, ``Deep learning for action recognition,'' in {\em Proceedings of the IEEE International Conference on Computer Vision (ICCV)}, 2014.

\bibitem{twostream1}
K.~Simonyan and A.~Zisserman, ``Two-stream convolutional networks for action recognition,'' in {\em Advances in Neural Information Processing Systems (NeurIPS)}, 2014.

\bibitem{twostream2}
S.~N. Gowda, ``Human activity recognition using combinatorial deep belief networks,'' in {\em Proceedings of the IEEE conference on computer vision and pattern recognition workshops}, pp.~1--6, 2017.

\bibitem{kinetics}
J.~Carreira and A.~Zisserman, ``Kinetics: A large-scale dataset for human action recognition,'' {\em Proceedings of the IEEE Conference on Computer Vision and Pattern Recognition (CVPR)}, 2017.

\bibitem{devise}
A.~Frome, G.~S. Corrado, J.~Shlens, S.~Bengio, J.~Dean, M.~Ranzato, and T.~Mikolov, ``Devise: A deep visual-semantic embedding model,'' {\em Advances in neural information processing systems}, vol.~26, 2013.

\bibitem{zsl_action_recognition1}
X.~Xu, T.~Hospedales, and S.~Gong, ``Semantic embedding space for zero-shot action recognition,'' in {\em 2015 IEEE International Conference on Image Processing (ICIP)}, pp.~63--67, IEEE, 2015.

\bibitem{zsl_action_recognition2}
J.~Qin, L.~Liu, L.~Shao, F.~Shen, B.~Ni, J.~Chen, and Y.~Wang, ``Zero-shot action recognition with error-correcting output codes,'' in {\em Proceedings of the IEEE Conference on Computer Vision and Pattern Recognition}, pp.~2833--2842, 2017.

\bibitem{truze}
S.~N. Gowda, L.~Sevilla-Lara, K.~Kim, F.~Keller, and M.~Rohrbach, ``A new split for evaluating true zero-shot action recognition,'' in {\em DAGM German Conference on Pattern Recognition}, pp.~191--205, Springer, 2021.

\bibitem{ahmad2023ez}
S.~Ahmad, S.~Chanda, and Y.~S. Rawat, ``Ez-clip: Efficient zeroshot video action recognition,'' {\em arXiv preprint arXiv:2312.08010}, 2023.

\bibitem{gowda2025temporal}
S.~Gowda, B.~Gao, X.~Gu, and X.~Jin, ``Is temporal prompting all we need for limited labeled action recognition?,'' in {\em Proceedings of the Computer Vision and Pattern Recognition Conference}, pp.~682--692, 2025.

\bibitem{I3D}
J.~Carreira and A.~Zisserman, ``Quo vadis, action recognition? a new model and the kinetics dataset,'' in {\em proceedings of the IEEE Conference on Computer Vision and Pattern Recognition}, pp.~6299--6308, 2017.

\bibitem{C3D}
C.~Li, Q.~Zhong, D.~Xie, and S.~Pu, ``Collaborative spatiotemporal feature learning for video action recognition,'' in {\em Proceedings of the ieee/cvf conference on computer vision and pattern recognition}, pp.~7872--7881, 2019.

\bibitem{bert}
J.~Devlin, ``Bert: Pre-training of deep bidirectional transformers for language understanding,'' {\em arXiv preprint arXiv:1810.04805}, 2018.

\bibitem{roberta}
Y.~Liu, ``Roberta: A robustly optimized bert pretraining approach,'' {\em arXiv preprint arXiv:1907.11692}, vol.~364, 2019.

\bibitem{od}
D.~Mandal, S.~Narayan, S.~K. Dwivedi, V.~Gupta, S.~Ahmed, F.~S. Khan, and L.~Shao, ``Out-of-distribution detection for generalized zero-shot action recognition,'' in {\em Proceedings of the IEEE Conference on Computer Vision and Pattern Recognition}, 2019.

\bibitem{gowda2023synthetic}
S.~N. Gowda, ``Synthetic sample selection for generalized zero-shot learning,'' in {\em Proceedings of the IEEE/CVF conference on computer vision and pattern recognition}, 2023.

\bibitem{ggm}
A.~Mishra, V.~K. Verma, M.~S.~K. Reddy, S.~Arulkumar, P.~Rai, and A.~Mittal, ``A generative approach to zero-shot and few-shot action recognition,'' in {\em 2018 IEEE Winter Conference on Applications of Computer Vision (WACV)}, 2018.

\bibitem{e2e}
B.~Brattoli, J.~Tighe, F.~Zhdanov, P.~Perona, and K.~Chalupka, ``Rethinking zero-shot video classification: End-to-end training for realistic applications,'' in {\em Proceedings of the IEEE/CVF Conference on Computer Vision and Pattern Recognition}, 2020.

\bibitem{er}
S.~Chen and D.~Huang, ``Elaborative rehearsal for zero-shot action recognition,'' in {\em Proceedings of the IEEE/CVF International Conference on Computer Vision}, 2021.

\bibitem{claster}
S.~N. Gowda, L.~Sevilla-Lara, F.~Keller, and M.~Rohrbach, ``Claster: clustering with reinforcement learning for zero-shot action recognition,'' in {\em European Conference on Computer Vision}, pp.~187--203, Springer, 2022.

\bibitem{SHANG2024112283}
J.~Shang, C.~Niu, X.~Tao, Z.~Zhou, and J.~Yang, ``Generalized zero-shot action recognition through reservation-based gate and semantic-enhanced contrastive learning,'' {\em Knowledge-Based Systems}, 2024.

\bibitem{niu2024superclass}
C.~Niu, J.~Shang, Z.~Zhou, and J.~Yang, ``Superclass-aware visual feature disentangling for generalized zero-shot learning,'' {\em Expert Systems with Applications}, 2024.

\bibitem{gil}
S.~N. Gowda, D.~Moltisanti, and L.~Sevilla-Lara, ``Continual learning improves zero-shot action recognition,'' in {\em Proceedings of the Asian Conference on Computer Vision}, pp.~3239--3256, 2024.

\end{thebibliography}

\end{document}